\documentclass{article}
\usepackage{ijcai19}

\usepackage{times}
\usepackage{soul}
\usepackage{url}
\usepackage[hidelinks]{hyperref}
\usepackage[utf8]{inputenc}
\usepackage[small]{caption}
\usepackage{graphicx}
\usepackage{amsmath}
\usepackage{booktabs}
\usepackage{algpseudocode}
\usepackage{todonotes}
\usepackage[linesnumbered,ruled,vlined]{algorithm2e}
\usepackage{latexsym} 
\usepackage{amssymb}
\usepackage{amsthm}

\newtheorem{theorem}{Theorem}
\newcommand{\norm}[1]{\left\lVert#1\right\rVert}
\newcommand{\el}{$\mathcal{EL}^{++}$}
\renewcommand{\Re}{\mathbb{R}}

\usepackage{cleveref}

\newlength\myindent
\title{EL Embeddings: Geometric construction of models for the Description Logic \el}

\author{
Maxat Kulmanov$^{1}$\and
Wang Liu-Wei$^1$\and
Yuan Yan$^{2}$\And
Robert Hoehndorf$^1$\footnote{Contact Author}\\
\affiliations
$^1$Computer, Electrical and Mathematical Sciences \&
    Engineering Division (CEMSE), Computational Bioscience Research
    Center (CBRC), King Abdullah University of Science and Technology,
    Thuwal 23955, Saudi Arabia\\
$^2$Department of Mathematics \& Statistics,
Dalhousie University, Halifax, Nova Scotia, Canada\\
\emails
\{maxat.kulmanov, liuwei.wang, robert.hoehndorf\}@kaust.edu.sa,
yuan.yan@dal.ca
}

\begin{document}

\maketitle

\begin{abstract}
  An embedding is a function that maps entities from one algebraic
  structure into another while preserving certain
  characteristics. Embeddings are being used successfully for mapping
  relational data or text into vector spaces where they can be used
  for machine learning, similarity search, or similar tasks. We
  address the problem of finding vector space embeddings for theories
  in the Description Logic \el that are also models of the TBox. To
  find such embeddings, we define an optimization problem that
  characterizes the model-theoretic semantics of the operators in \el
  within $\Re^n$, thereby solving the problem of finding an
  interpretation function for an \el theory given a particular domain
  $\Delta$. Our approach is mainly relevant to large \el theories and
  knowledge bases such as the ontologies and knowledge graphs used in
  the life sciences. We demonstrate that our method can be used for
  improved prediction of protein--protein interactions when compared
  to semantic similarity measures or knowledge graph embeddings.
\end{abstract}

\section{Introduction}

There has been a recent proliferation of methods that generate
``embeddings'' for different types of entities. Often, these
embeddings are functions that map entities within a certain structure
into a vector space $\Re^n$ such that a set of structural
characteristics of the original structure are preserved within the
vector space. For example, word embeddings are generated for words
within a corpus of text based on the distribution of words and their
co-mentions \cite{word2vec}.

Knowledge graph embeddings are used to project sets of discrete facts
into a vector space over real numbers and aim to preserve some
structural properties of the graph within $\Re^n$
\cite{Nickel2016review,Wang2017}.
% Given a knowledge graph $KG = (...)$, a knowledge graph embedding is a
% function $f_\eta: KG \mapsto \Re^n$ for some parameter $n$.
The embeddings can project entities and relations within a knowledge
graph into $\Re^n$ such that they can naturally be used as features
for machine learning tasks such as classification, regression, or
clustering, or directly utilize similarity measures within $\Re^n$ for
determining semantic similarity, performing reasoning by analogy, and
thereby predict relations. Most knowledge graph embedding approaches
find the embedding function through optimization with respect to an
objective function and, optionally, a set of constraints
\cite{Wang2017}.

Inference in knowledge graphs is often limited to composition of
relations. Model-theoretic languages such as Description Logics can
also be used to express relational knowledge while adding operators
that cannot easily be expressed in graph-based form (quantifiers,
negation, conjunction, disjunction) \cite{Baader2003}. In particular
the life sciences have developed a large number of ontologies
formulated in the Web Ontology Language (OWL) \cite{Grau2008}, and
many of the life science ontologies fall in the OWL 2 EL profile
\cite{elvira}, which is based on the Description Logics \el
\cite{owlprofiles}. The life science ontologies are used to express
domain knowledge and serve as a foundation for analysis and
interpretation of biological data, for example through statistical
measures \cite{Subramanian2005} or semantic similarity measures
\cite{Couto2009}. Recently, ``ontology embeddings'' were developed for
life science ontologies that map classes, relations, and instances in
these ontologies into a vector space while preserving certain
syntactic properties of the ontology axioms and their deductive
closure \cite{opa2vec}. However, embeddings that rely primarily on
preserving syntactic properties of knowledge bases within a vector
space are limited by the kind of inferences that can be precomputed
and expressed in the knowledge representation language, and do not
utilize prior knowledge about the semantics of operators during the
search for an embedding function.

Here, we introduce EL Embeddings, a method to generate embeddings for
ontologies in the Description Logic \el. EL Embeddings explicitly
generate -- or approximate -- models for an \el theory and therefore
approximate the interpretation function. % For
% first-order languages, the existence of models in $\Re^n$ is
% guaranteed if there is an infinite model.
For this purpose, we formulate the problem of finding a model as an
optimization problem over $\Re^n$. An alternative view on EL
Embeddings is that we extend knowledge graph embeddings with the
semantics of conjunction, existential quantification, and the bottom
concept.

We demonstrate that the resulting embeddings can be used for
determining semantic similarity or suggest axioms that may be entailed
by the theory.  As large \el theories are mainly used in the life
sciences, we evaluate our approach on a large knowledge base of
protein--protein interactions and protein functions. We show that our
method can improve the prediction of protein--protein interactions
when compared to semantic similarity measures and to knowledge graph
embeddings.

\section{Related work}
\subsection{The Description Logic \el and its application
  in life sciences}

\begin{table}
  \centering
  \begin{tabular}{|p{1.9cm}|c|p{3.7cm}|}
    \hline
    {\bf} Name & Syntax & Semantics \\
    \hline
    top & $\top$ & $\Delta^{\mathcal{I}}$ \\
    \hline
    bottom & $\bot$ & $\emptyset$ \\
    \hline
    nominal & $\{ a \} $ & $\{ a^{\mathcal{I}} \}$ \\
    \hline
    conjunction & $C \sqcap D$ & $ C^{\mathcal{I}} \cap
                                 D^{\mathcal{I}}$ \\
    \hline
    existential restriction & $\exists r.C$ & $ \{ x \in
                                              \Delta^{\mathcal{I}} |
                                              \exists y \in
                                              \Delta^{\mathcal{I}} :
                                              (x,y) \in
                                              r^{\mathcal{I}} \land y
                                              \in C^{\mathcal{I}} \} $
    \\
    \hline
    generalized concept inclusion & $C \sqsubseteq D$ &
                                                        $C^{\mathcal{I}}
                                                        \subseteq
                                                        D^{\mathcal{I}}$
    \\
    \hline
    instantiation & $C(a)$ & $a^\mathcal{I} \in C^\mathcal{I}$ \\
    \hline
    role assertion & $r(a,b)$ & $(a^\mathcal{I},b^\mathcal{I}) \in r^\mathcal{I}$ \\
    \hline
    % role inclusion & $r_1 \circ ... \circ r_n \sqsubseteq r$ &
    %                                                            $r_1^{\mathcal{I}}
    %                                                            \circ
    %                                                            ... \circ
    %                                                            r_n^{\mathcal{I}}
    %                                                            \subseteq
    %                                                            r^{\mathcal{I}}$
    %    \\
    % \hline
    
  \end{tabular}
  \caption{\label{tbl:el}Syntax and semantic of $\mathcal{EL}^{++}$
    (omitting role inclusions and concrete domains).}
\end{table}

The Description Logic $\mathcal{EL}^{++}$
\cite{el-report} is a Description Logic for which
subsumption can be decided in polynomial time and which is therefore
suitable for representing and reasoning over large ontologies. The
syntax and semantics of $\mathcal{EL}^{++}$ is summarized in Table
\ref{tbl:el} (omitting concrete domains which we will not consider
here). \el also forms the basis of the OWL 2 EL profile of OWL
\cite{owlprofiles}.

The ABox axioms (instantiation and role assertion) in \el can be
eliminated by replacing $C(a)$ with $\{ a \} \sqsubseteq C$ and
$r(a,b)$ with $\{ a \} \sqsubseteq \exists r.\{ b \}$, and every \el
TBox can be normalized into one of four normal forms:
$C \sqsubseteq D$, $C \sqcap D \sqsubseteq E$,
$\exists R.C \sqsubseteq D$, and $C \sqsubseteq \exists R.D$ (where
the bottom concept can only appear on the right-hand side and only in
the first three normal forms)
\cite{el-report}.
% First, the ABox is eliminated by replacing each individual $a$ with
% the singleton class $\{ a \}$ and each role assertion $r(a,b)$ with
% the TBox axiom $\{ a \} \sqsubseteq \exists r.\{ b \}$.

$\mathcal{EL}^{++}$ is widely used to represent and reason over life
science ontologies such as the Gene Ontology \cite{Ashburner2000},
the Human Phenotype Ontology \cite{hpo}, or SNOMED CT
\cite{Schulz2009}. These ontologies are often large and require fast
decision procedures for automated reasoning, which \el can provide
\cite{el-report}. The ontologies in the life-science domain are also
used as components in knowledge graphs to structure data and provide
background knowledge about classes within their domains.

% defined as: let $N_C$ and $N_R$ be two disjoint
% sets of symbols, the {\em concept names} and {\em role names},
% respectively. An $\mathcal{EL}^{++}$ concept is inductively defined

\subsection{Knowledge graph embeddings}

% \todo[inline]{Chase: 
% Inducing logic rules
% Applications in life sciences
% Ontologies and knowledge graphs
% ManifoldE for ``precise'' link prediction.}
Knowledge graph embedding methods have been developed to map entities
and their relations expressed in a knowledge graph into a vector space
while preserving relational and other semantic information under
certain vector space relations
\cite{Nickel2016review}. Translation-based embeddings, such as TransE
\cite{transe}, generate vector space representations of entities and
relations in a graph such that
$ \mathbf{a} + \mathbf{r} \approx \mathbf{b}$ if $r(a,b)$ is a
relation in the knowledge graph. Other approaches include methods for
exploring the neighborhood of nodes in the graph and encoding these
nodes and their relations \cite{Wang2017}.
% by
% based on simple vector arithmetic
% relations between entities and relations. Random walk based methods,
% inspired by DeepWalk \cite{deepwalk}, exploits local neighborhood
% features.

Knowledge graphs are heterogeneous graphs with an explicit semantics
and an inference relation; one way in which the semantics of relations
in a knowledge graph can be taken into account when generating
knowledge graph embeddings is by pre-computing a limited form of
deductive closure on the graph before finding the embeddings
\cite{Nickel2016review,Wang2017}. Such an approach has also been
applied successfully in the life sciences where knowledge graph
embeddings based on deductively closed graphs have been used for
predicting gene--disease associations or drug targets \cite{neurosym}.

\subsection{Semantic similarity}

A related yet alternative approach to using knowledge graph embeddings
for relational learning is the use of semantic similarity measures to
compare two classes within an ontology, or two instances with respect
to the axioms within an ontology \cite{Couto2009}. There is a wide
range of semantic similarity measures, most of which operate on graphs
or sets constructed from a theory syntactically (e.g., by applying a
certain closure on a theory to generate graphs) but can
also be applied to model structures such as the canonical models of
$\mathcal{ALC}$ theories \cite{harispe15}.

In life sciences, semantic similarity measures can be applied
predictively \cite{Couto2009}; ontologies provide biological features,
and similarity between the biological features can be indicative of an
underlying biological relation. For example, semantic similarity
between proteins linked to functions in the Gene Ontology
\cite{Ashburner2000} can be used to determine protein--protein
interactions based on the biological assumption that interacting
proteins are likely to have similar functions
\cite{Kulmanov2017}; similarly, semantic similarity
measures are used to identify candidate genes associated with diseases
\cite{funsimat}. Widely-applied semantic similarity measures
in life sciences include Resnik's similarity \cite{Resnik1995} or the
weighted Jaccard index \cite{Couto2009}. Recently, semantic similarity
is also measured based on knowledge graph embeddings, for example for
predicting protein--protein interactions \cite{opa2vec}.

\section{Geometric models for $\mathcal{EL}^{++}$}

\subsection{Relation model and normalization}
Our aim is to extend knowledge graph embeddings so that they
incorporate the \el operators (conjunction, existential
quantification) and can express the bottom concept $\bot$. We use a
relational embedding model, TransE \cite{transe}, to map relations into
$\Re^n$. We chose TransE due to its simplicity; however, our method
can accommodate different relational models.

Let $O = (\mathbb{C}, \mathbb{R}, \mathbb{I}; ax)$ be an \el ontology
consisting of a set of class symbols $\mathbb{C}$, relation symbols
$\mathbb{R}$, individual symbols $\mathbb{I}$, and a set of axioms
$ax$. We first transform $ax$ into a normal form following \cite{el-report}; we
eliminate the ABox by replacing each individual symbol with a
singleton class and rewriting relation assertions $r(a,b)$ and class
assertions $C(a)$ as $\{ a \} \sqsubseteq \exists r.\{ b \}$ and
$\{ a \} \sqsubseteq C$. Using the conversion rules in \cite{el-report} we
transform the set of axioms into one of four forms where $C, D, E \in
\mathbb{C}$ and $R \in \mathbb{R}$: $C \sqsubseteq D$; $C \sqcap D
\sqsubseteq E$; $C \sqsubseteq \exists R.D$; $\exists R.C \sqsubseteq
D$.

\subsection{Objective functions}

If an \el theory $T$ has a model then it also has an infinite model,
and therefore it also has a model with a universe of $\Re^n$ for any
$n$ (L\"{o}wenheim--Skolem upwards) \cite{lpl}.  The embedding
function our model aims to find is intended to approximate the
interpretation function $\mathcal{I}$ in the \el semantics (Table
\ref{tbl:el}).  Specifically, our embedding function $\eta$ aims to map each
class $C$ to an open $n$-ball in $\Re^n$ ($\eta(C)$) and every binary
relation $r$ to a vector in $\Re^n$. 
We define a geometric ontology embedding $\eta$ as a pair $(f_\eta,
r_\eta)$ of functions that map classes and relations in $O$ into
$\Re^n$, $f_\eta : C \cup R \mapsto \Re^n$ and $r_\eta : C \mapsto
\Re$. The function $f_\eta(x)$ maps a class to its center or maps a
relation to its embedding vector, and $r_\eta(x)$ maps a class $x$ to
the radius associated with it.

We formulate one loss function for each of the normal forms so that
the embedding $\eta$ preserves the semantics of \el geometrical within
$\Re^n$. The total loss for finding $\eta$ is the sum of the loss
functions for all the normal forms.  We first assume that none of the
classes are $\bot$. The first loss function (Eqn. \ref{eqn:nf1}) aims
to capture the notion that, if $C \sqsubseteq D$, then $\eta(C)$
should lie in $\eta(D)$. For all loss functions we use a margin
parameter $\gamma$; if $\gamma < 0$ then $\eta(C)$ lies properly
inside $\eta(D)$. Also, we add normalization loss for all class
embeddings in the loss functions, essentially moving the
centers of all $n$-balls representing classes to lie on the unity
sphere.

\begin{equation}
  \label{eqn:nf1}
\begin{split}
  & loss_{C \sqsubseteq D}(c,d) = \\
  & \max(0, \norm{f_\eta(c) - f_\eta(d)} + r_\eta(c) - r_\eta(d) - \gamma) \\
  & + |\norm{f_\eta(c)} - 1| + |\norm{f_\eta(d)} - 1|
\end{split}
\end{equation}

The loss function for the second normal form (Eqn. \ref{eqn:nf2}),
$C \sqcap D \sqsubseteq E$, should capture the notion that the
intersection or overlap of the $n$-balls representing $C$ and $D$
should lie within the $n$-ball representing $E$; while the overlap
between $\eta(C)$ and $\eta(D)$ is not in general an $n$-ball, the loss
should characterize the smallest $n$-ball which includes the intersection of
$\eta(C)$ and $\eta(D)$ and minimizes its non-overlap with
$\eta(E)$.
% The center and radius of the smallest $n$-ball containing the intersection of
% $\eta(C)$ and $\eta(D)$ are $f_\eta(c)+\frac{r_\eta(c)^2-r_\eta(d)^2+\norm{f_\eta(c) - f_\eta(d)}^2}{2\norm{f_\eta(c) - f_\eta(d)}^2}(f_\eta(d)-f_\eta(c))$, $\frac{\sqrt{4\norm{f_\eta(c) - f_\eta(d)}^2r_\eta(c)^2-(r_\eta(c)^2-r_\eta(d)^2+\norm{f_\eta(c) - f_\eta(d)}^2)^2}}{2\norm{f_\eta(c) - f_\eta(d)}} $, respectively.
Let
$h=\frac{r_\eta(c)^2-r_\eta(d)^2+\norm{f_\eta(c) -
    f_\eta(d)}^2}{2\norm{f_\eta(c) - f_\eta(d)}}$, then the center and
radius of the smallest $n$-ball containing the intersection of
$\eta(C)$ and $\eta(D)$ are
$f_\eta(c)+\frac{h}{\norm{f_\eta(c) -
    f_\eta(d)}}(f_\eta(d)-f_\eta(c))$ and $\sqrt{r_\eta(c)^2-h^2}$,
respectively.  However, we found it difficult to implement this loss
due to very large gradients and therefore use the approximation of this loss
given in Eqn. \ref{eqn:nf2}.  The first term in Eqn. \ref{eqn:nf2} is
a penalty when the $n$-balls representing $C$ and $D$ are disjoint;
the second and third terms force the center of $\eta(E)$ to lie inside
the intersection of $\eta(C)$ and $\eta(D)$; the fourth term makes the radius of
$\eta(E)$ to be larger than the radius of the smallest $n$-balls of
the intersecting classes; this radius is strictly larger than the
radius of the smallest $n$-ball containing the intersection and
therefore satisfies the condition that the intersection should lie
within $\eta(E)$.
\begin{equation}
  \label{eqn:nf2}
\begin{split} 
&loss_{C \sqcap D \sqsubseteq E}(c,d,e) = \\
    & \max(0, \norm{f_\eta(c) - f_\eta(d)} - r_\eta(c) - r_\eta(d) - \gamma) \\
    & + \max(0, \norm{f_\eta(c) - f_\eta(e)} - r_\eta(c) - \gamma) \\
    & + \max(0, \norm{f_\eta(d) - f_\eta(e)} - r_\eta(c) - \gamma) \\
    & +  \max(0, \min(r_\eta(c), r_\eta(d)) - r_\eta(e) - \gamma)\\
    & + |\norm{f_\eta(c)} - 1| + |\norm{f_\eta(d)} - 1| + |\norm{f_\eta(e)} - 1|
  \end{split}
\end{equation}

The first two normal forms do not include any quantifiers or
relations. Every point that lies properly within an $n$-ball
representing a class is a potential instance of that class, and we
apply relations as transformations on these points (following the
TransE relation model). Therefore, relations are transformations on
$n$-balls. Equations \ref{eqn:nf3} and \ref{eqn:nf4} capture this
intention. 
\begin{equation}
  \label{eqn:nf3}
\begin{split}
& loss_{C \sqsubseteq \exists R.D}(c,d,r) = \\
   & \max(0, \norm{f_\eta(c) + f_\eta(r) - f_\eta(d)}  + r_\eta(c) - r_\eta(d) - \gamma) \\
  & + |\norm{f_\eta(c)} - 1| + |\norm{f_\eta(d)} - 1|
  \end{split}
\end{equation}

\begin{equation}
  \label{eqn:nf4}
  \begin{split}
  & loss_{\exists R.C \sqsubseteq D}(c,d,r) = \\
  & \max(0, \norm{f_\eta(c) - f_\eta(r) - f_\eta(d)} - r_\eta(c) - r_\eta(d) - \gamma) \\
  & + |\norm{f_\eta(c)} - 1| + |\norm{f_\eta(d)} - 1|
\end{split}
\end{equation}

In the normal forms for \el, $\bot$ can only occur on the right-hand
side in three of the normal forms. We formulate separate loss
functions for the cases in which $\bot$ appear. First, $C \sqcap D
\sqsubseteq \bot$ states that $C$ and $D$ are disjoint and therefore
$\eta(C)$ and $\eta(D)$ should not overlap. Equation
\ref{eqn:disjoint} captures disjointness loss.
\begin{equation}
  \label{eqn:disjoint}
\begin{split} 
&loss_{C \sqcap D \sqsubseteq \bot}(c,d,e) = \\
    & \max(0, r_\eta(c) + r_\eta(d) - \norm{f_\eta(c) - f_\eta(d)} + \gamma) \\
    & + |\norm{f_\eta(c)} - 1| + |\norm{f_\eta(d)} - 1|
  \end{split}
\end{equation}

Loss \ref{eqn:empty} captures the intuition that a class is
unsatisfiable by minimizing the radius $r_\eta$ of the class. 
\begin{equation}
  \label{eqn:empty}
\begin{split}
  loss_{C \sqsubseteq \bot}(c) = r_\eta(c)
\end{split}
\end{equation}

Since we use TransE as model for relations, the radius of an $n$-ball
cannot change after a transformation by a relation. Therefore, we use
the same loss (Eqn. \ref{eqn:empty2}) for $\exists R.C \sqsubseteq
\bot$.
\begin{equation}
  \label{eqn:empty2}
  \begin{split}
  loss_{\exists R.C \sqsubseteq \bot}(c,r) = r_\eta(c)
  \end{split}
\end{equation}

While our model does not need negatives, we can use negatives as in
translating embeddings to improve predictive performance. For this
purpose, we add an optional loss for $C \not\sqsubseteq \exists R.D$
as in Equation \ref{eqn:negatives}.
\begin{equation}
  \label{eqn:negatives}
\begin{split}
& loss_{C \not\sqsubseteq \exists R.D}(c,d,r) = \\
   & \max(0, r_\eta(c) + r_\eta(d) - \norm{f_\eta(c) + f_\eta(r) - f_\eta(d)} + \gamma) \\
  & + |\norm{f_\eta(c)} - 1| + |\norm{f_\eta(d)} - 1|
  \end{split}
\end{equation}

% \begin{equation}
%   \begin{split}
%   loss_{\top}(\top) = |r_\eta(\top) - \infty|
%   \end{split}
% \end{equation}

Finally, we add the constraints $r_\eta(\top) = \infty$ to capture the
intuition that the interpretation of $\top$ is $\Delta^\mathcal{I} =
\Re^n$, and $r_\eta(x) \geq 0$ for all $x$. 

\subsection{Embeddings and models}

\begin{theorem}[Correctness]
  Let $T$ be a theory in \el. If $\gamma \leq 0$ and
  $loss_n(\eta(T)) = 0$ then $T$ has a model.
  % $\mathcal{I} = (\Re^n, f_\eta^{\mathcal{I}})$.
  % The loss functions approximate the interpretation function
  % $\mathcal{I}$. If loss = 0, $f_\eta \in \mathcal{I}$ with $\Delta =
  % \Re^n$.
\end{theorem}
We outline a proof of this theorem.
First, we set $\Delta = \Re^n$.
By construction, $\top = \Delta^\mathcal{I} = \Re^n$. We interpret
each class $C$ as the set of points lying within the open $n$-ball
$\eta(C)$, $C^{\mathcal{I}} = \{ x \in \Re^n | \norm{f_\eta(C) - x} <
r_\eta(C) \}$ and every binary relation $r$ as a set of tuples
$r^\mathcal{I} = \{ (x,y) | x + f_\eta(r) = y \}$.
We need to show that the conditions in Table \ref{tbl:el} are satisfied if
$loss(T) = 0$. The loss is the sum of losses for the four normal
forms, all of which are non-negative.

The remaining conditions are preserved for each of the four normal
forms and their respective losses: by construction, normal form 1
ensures that, if $C \sqsubseteq D$ is in the TBox, then
$C^\mathcal{I} \subseteq D^\mathcal{I}$; the loss of normal form 2,
$C \sqcap D \sqsubseteq E$, constructs the smallest $n$-ball containing
the intersection of $\eta(C)$ and $\eta(D)$ and ensures that this $n$-ball lies
within the $\eta(E)$; normal form 3, $C \sqsubseteq \exists R.D$,
applies a relation transformation to all instances $x$ of $C$ (i.e.,
it constructs $f_\eta(x) + f_\eta(R)$ for all elements $x$ of the
$n$-ball $\eta(C)$) and ensures that each instance of $C$ lies within
$\eta(D)$, therefore ensuring that
$\{ x \in \Re^n | \norm{f_\eta(C) - x} < r_\eta(C)\} \subseteq \{ x
\in \Re^n | \norm{f_\eta(D) + f_\eta(R) - x } < r_\eta(D) \} $ and
therefore
$\eta(C) \subseteq \{ x \in \Re^n | \exists y \in \Delta^{\mathcal{I}}
: (x,y) \in R^{\mathcal{I}} \land y \in D^{\mathcal{I}} \}$; normal
form 4 trivially satisfies $\exists R.C \sqsubseteq D$. It follows
similarly from the loss functions \ref{eqn:disjoint}--\ref{eqn:empty2}
that $\bot$ is interpreted as $\emptyset$; the only case requiring
more attention is $C \sqcap D \sqsubseteq \bot$ where it is possible
that the hyperspheres bounding the $n$-balls $\eta(C)$ and $\eta(D)$
touch. In our interpretation, we assume that hyperballs are open so
that the $n$-balls are disjoint even if their bounding hyperspheres
touch.

\subsection{Training and Implementation}

While our algorithm can find, or approximate, a model without any
negative samples for any of the four normal forms, these models are
usually underspecified. We intend to use our embeddings for relational
learning which benefits from a representation in which asserted and
implied axioms can be discriminated from those that should not hold
true. Therefore, we follow a similar strategy for sampling negatives
as in TransE and randomly generate corrupted axioms in third normal
form ($C \sqsubseteq \exists R.D$) by replacing either $C$ or $D$ with
a class $C'$ or $D'$ such that neither $C' \sqsubseteq \exists R.D$
nor $C \sqsubseteq \exists R.D'$ are asserted axioms in the ontology.

We randomly initialize the embeddings for classes and relations. We
then sample formulas for each loss function in mini-batches and update
the embeddings with respect to the sum of the loss functions (see
Algorithm \ref{algorithm}). We implement the algorithm in two
parts. First, the processing of ontologies in OWL format and
normalization into the \el normal forms are performed using the OWL API
and the APIs provided by the jCel reasoner which implements the \el
normalization rules \cite{jcel}. Training of embeddings and
optimization is done using Python and the TensorFlow library, and we
use the Adam optimizer \cite{adam} for updating embeddings.

\begin{algorithm}
  \SetAlgoLined
  \DontPrintSemicolon
  \SetKwData{Left}{left}\SetKwData{This}{this}\SetKwData{Up}{up}
  \SetKwFunction{Union}{Union}\SetKwFunction{FindCompress}{FindCompress}
  \SetKwInOut{Input}{input}\SetKwInOut{Output}{output}
  \Input{An ontology $O = (C, R, I; ax)$ in OWL format; margin
    $\gamma$; dimension $n$; epochs $epochs$; batchsize $bs$}
  \Output{embeddings $(f_\eta, r_\eta)$}
  \BlankLine
  // \emph{eliminate ABox} \;
%  \For{$a \in I$}{
  $C \leftarrow C \cup \{ a \} $ for each $a \in I$\;
%  }
%  \For{$r(a,b) \in ax$}{
  $ax \leftarrow ax \cup (\{ a \} \sqsubseteq \exists r.\{ b \}) $
  for each $r(a,b) \in ax$\;
%  }
%  \For{$C(a) \in ax$}{
  $ax \leftarrow ax \cup (\{ a \} \sqsubseteq C) $ for each $C(a)
  \in ax$ \;
%  }
  // \emph{apply \el normalization rules} \;
  $(ax_{nf1}, ax_{nf2}, ax_{nf3}, ax_{nf4}) = normalize(ax)$ \; 
  // \emph{separate axioms with $\bot$} for NF 1, 2 and 4\; 
  $(ax_{nf1}, ax_{bot1}) = separate(ax_{nf1}) $\;
  $(ax_{nf2}, ax_{bot2}) = separate(ax_{nf3}) $\;
  $(ax_{nf4}, ax_{bot4}) = separate(ax_{nf4}) $\;
  // \emph{generate negatives for proteins in NF 3} \;
  $neg_{nf3} = negatives(ax_{nf3})$ \;
  $D \leftarrow \{ax_{nf1}\} \cup \{ax_{nf2}\} \cup \{ax_{nf3}\} \cup \{ax_{nf4}\} \cup \{ax_{bot1}\} \cup \{ax_{bot2}\} \cup \{ax_{bot4}\} \cup \{neg_{nf3}\}$ \;
  
  // {\em initialize embeddings } \;
  $f_\eta(c) = uniform(0, 1)$ for each $c \in C$ \;
  $r_\eta(c) = uniform(0, 1)$ for each $c \in C$ \;
  $f_\eta(r) = uniform(0, 1)$ for each $r \in R$ \;
  
  % // {\em normalize embeddings} \;
  % $f_\eta(c) = \frac{f_\eta(c)}{\norm{f_\eta(c)}}$ for each $c \in C$ \;
  % $f_\eta(r) = \frac{f_\eta(r)}{\norm{f_\eta(r)}}$ for each $r \in R$ \;
 
  % Optimization/Training:
  \For{$e \in epochs$} {
  // \emph{Randomly sample a minibatch of size bs for each loss type} \;
  $(s_{nf1}, s_{nf2}, s_{nf3}, s_{nf4}, s_{bot1}, s_{bot2}, s_{bot4}, s_{neg}) = sample(D, bs)$ \;
  // \emph{Update embeddings w.r.t.} \;
  $\sum \nabla loss(s_{nf1}, s_{nf2}, s_{nf3}, s_{nf4}, s_{bot1}, s_{bot2}, s_{bot4}, s_{neg}) $
  }
  \caption{\label{algorithm}Algorithm used for training EL Embeddings.}
\end{algorithm}

\begin{table*}[ht]
  \begin{tabular}{|p{1.7cm}|p{1.7cm}|p{1.8cm}|p{1.7cm}|p{1.8cm}|p{1.5cm}|p{1.5cm}|p{1cm}|p{1.5cm}|}
    \hline
    Method & Raw Hits@10 & Filtered Hits@10 & Raw Hits@100 & Filtered Hits@100 & Raw Mean Rank & Filtered Mean Rank & Raw AUC & Filtered AUC \\ 
    \hline
    TransE (RDF) & 0.03 & 0.05 & 0.22 & 0.27 & 855 & 809 & 0.84 & 0.85 \\
    \hline
    TransE (plain) & 0.06 & 0.13 & 0.41 & 0.54 & 378 & 330 & 0.93 & 0.94 \\
    \hline
    SimResnik & 0.08 & 0.18 & 0.38 & 0.49 & 713 & 663 & 0.87 & 0.88 \\
    \hline
    SimLin & 0.08 & 0.17 & 0.34 & 0.45 & 807 & 756 & 0.85 & 0.86 \\
    \hline
    EL Embeddings & \textbf{0.10} & \textbf{0.23} & \textbf{0.50} & \textbf{0.75} & \textbf{247} & \textbf{187} & \textbf{0.96} & \textbf{0.97} \\
    \hline
  \end{tabular}
  \caption{\label{tbl:results-yeast} Prediction performance for yeast
    protein--protein interactions.}
\end{table*}

\begin{table*}[ht]
  \begin{tabular}{|p{1.5cm}|p{1.7cm}|p{1.8cm}|p{1.7cm}|p{1.8cm}|p{1.5cm}|p{1.5cm}|p{1cm}|p{1.5cm}|}
    \hline
    Method & Raw Hits@10 & Filtered Hits@10 & Raw Hits@100 & Filtered Hits@100 & Raw Mean Rank & Filtered Mean Rank & Raw AUC & Filtered AUC \\ 
    \hline
    TransE (RDF)& 0.02 & 0.03 & 0.12 & 0.16 & 2262 & 2189 & 0.85 & 0.85 \\
    \hline
    TransE (plain)& 0.05 & 0.11 & 0.32 & 0.44 & 809 & 737 & \textbf{0.95} & 0.95 \\
    \hline
    SimResnik  & 0.05 & 0.10 & 0.23 & 0.28 & 2549 & 2475 & 0.83 & 0.83 \\
    \hline
    SimLin & 0.04 & 0.08 & 0.19 & 0.22 & 2818 & 2743 & 0.81 & 0.82 \\
    \hline
    EL Embeddings & \textbf{0.09} & \textbf{0.22} & \textbf{0.43} & \textbf{0.70} & \textbf{707} & \textbf{622} & \textbf{0.95} & \textbf{0.96} \\
    \hline
  \end{tabular}
  \caption{\label{tbl:results-human} Prediction performance for human
    protein--protein interactions.}
\end{table*}

\section{Experiments}

\subsection{Example: Family domain}
\begin{figure}
    \centering
    \includegraphics[width=0.45\textwidth]{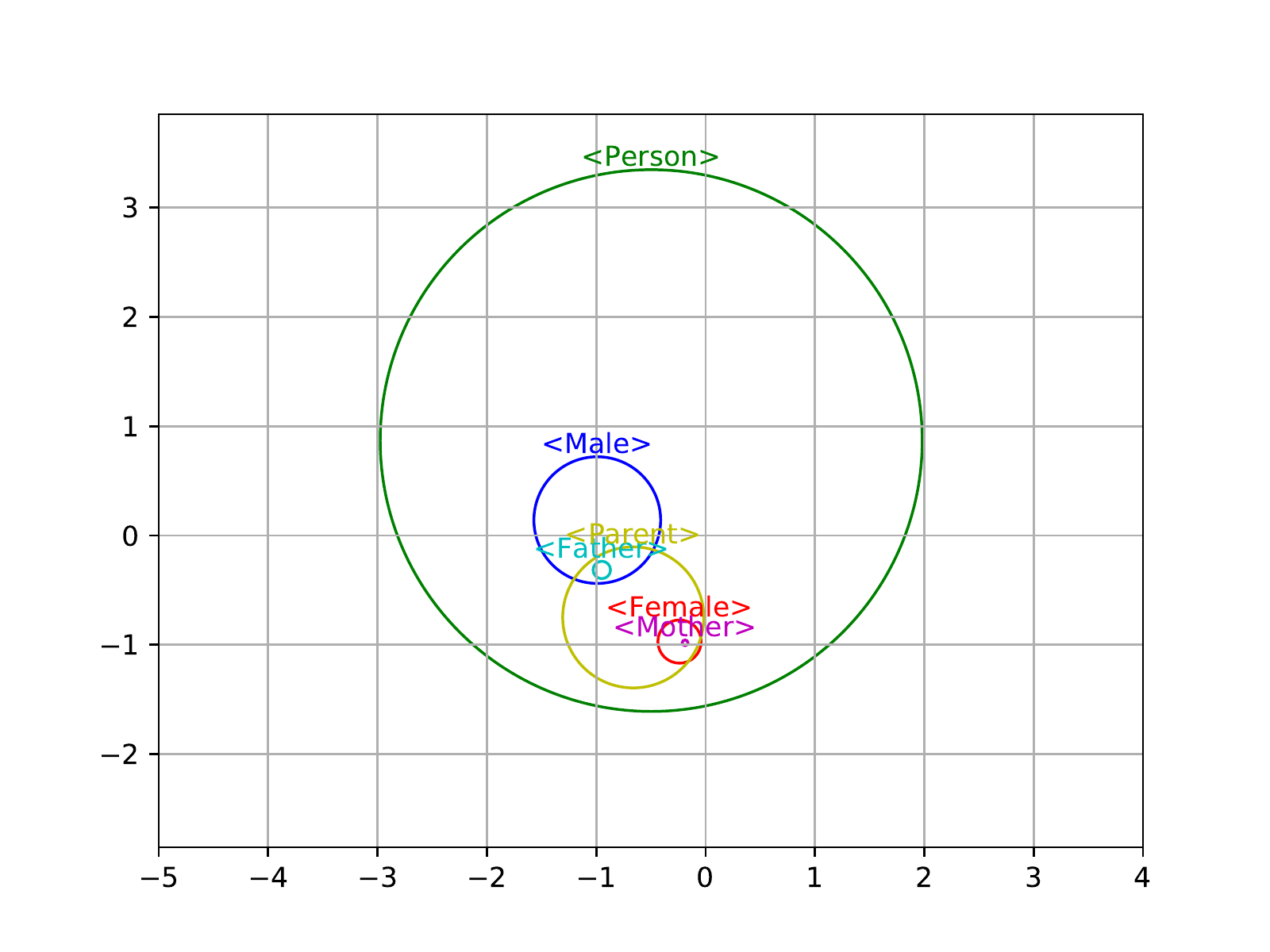}
    \caption{Visualization of embeddings in the family domain example.}
    \label{fig:embeds}
\end{figure}

We first construct a simple test knowledge base to test our model. We
use the family domain in which we generate a knowledge base that
contains examples for each of the normal forms
(Eqn. \ref{fam1}--\ref{famlast}). We chose a margin $\gamma = 0$ and
an embedding dimension of $2$ so that we can visualize the generated
embeddings in $\Re^2$. Figure \ref{fig:embeds} shows the resulting
embeddings. 
\begin{eqnarray}
  \label{fam1}
& Male & \sqsubseteq Person \\
& Female & \sqsubseteq Person \\
& Father & \sqsubseteq Male \\
& Mother & \sqsubseteq Female \\
& Father & \sqsubseteq Parent \\
& Mother & \sqsubseteq Parent \\
& Female \sqcap Male & \sqsubseteq \bot \\
& Female \sqcap Parent & \sqsubseteq Mother \\
& Male \sqcap Parent & \sqsubseteq Father \\
& \exists hasChild.Person & \sqsubseteq Parent \\
& Parent & \sqsubseteq Person \\
& Parent & \sqsubseteq \exists hasChild.\top
  \label{famlast}
%& Person & \sqsubseteq \top
\end{eqnarray}

\subsection{Protein--protein interactions}

Prediction of interactions between proteins is a common task in
molecular biology that relies on information about sequences as well
as functional information \cite{Couto2009,Kulmanov2017}. The
information about the functions of proteins is represented through the
Gene Ontology (GO) \cite{Ashburner2000}, a large manually-created
ontology with over 45,000 classes and 100,000 axioms. GO can be
formalized in OWL 2 EL and therefore falls in the \el formalism
\cite{Golbreich2007}. Common approaches to predicting protein--protein
interactions (PPIs) include network-based approaches and the use of
semantic similarity measures \cite{Kulmanov2017}.

We use the PPI dataset provided by the STRING database
\cite{string105} to construct a knowledge graph of proteins and their
interactions. We construct two graphs for human and yeast organisms
with relations for which a confidence score of 700 or more is assigned
in STRING (following recommendations in STRING
\cite{string105}); if an interaction between two proteins
$P_1$ and $P_2$ exists in STRING, we assert $interacts(P_1,P_2)$. We
further add the associations of proteins with functions from the GO,
provided by STRING, together with all classes and relations from
GO. For this we use two representation patterns. First, we generate an
OWL representation in which proteins are instances, and if protein
$P$ is associated with the function $F$ we add the axiom
$\{ P \} \sqsubseteq \exists hasFunction.F$ (based on the ABox axiom
$(\exists hasFunction.F)(P)$). We use this information together with
the native OWL version of GO provided by the OBO Foundry repository
\cite{Smith2007}; when applying knowledge graph embeddings to this
representation, we use the RDF serialization of the complete OWL
knowledge base as the knowledge graph. While the OWL-based
representation is suitable for our EL Embeddings, knowledge graph
embeddings and semantic similarity measures would benefit from a
graph-based representation. We therefore create a second
representation in which we replace all axioms of the type
$X \sqsubseteq \exists R.Y$ in GO with a relation $R(X,Y)$, and link
proteins to their functions using a $hasFunction$ relation (i.e., if
protein $P$ has function $F$, we assert $R(P,F)$).

We generate a training, testing, and validation split (80\%/10\%/10\%)
from interaction pairs of proteins.  We use the TransE \cite{transe}
implementation in the PyKEEN framework \cite{pykeen} on both
representations (native OWL/RDF and the ``plain'' representation) to
generate knowledge graph embeddings and use them for link
prediction. We implement two semantic similarity measures, Resnik's
similarity \cite{Resnik1995} and Lin's similarity \cite{harispe15},
together with the best-match average strategy for combining pairwise
class similarities \cite{Couto2009,harispe15}, and compute the
similarity between proteins based on their associations with GO
classes.  To predict PPIs with EL Embeddings, we predict whether
axioms of the type $\{ P \} \sqsubseteq \exists hasFunction.\{ F \}$
hold. We use the similarity-based function in Eqn. \ref{eqn:predict}
for this prediction.
\begin{equation}
  \label{eqn:predict}
  \begin{split}
   sim(c, r, d) =  -\max&(0, \norm{f_\eta(c) + f_\eta(r) -
     f_\eta(d)} \\
   - r_\eta(c) & - r_\eta(d) - \gamma)
  \end{split}
\end{equation}

We evaluate the predictive performance based on recall at rank $10$,
rank $100$, mean rank and area under the ROC curve using our
validation set. In our experiments, we perform an extensive search for
optimal parameters for TransE and our EL Embeddings, testing
embeddings sizes of $50$, $100$, $200$, and $400$. We also evaluate
the performance with different margin parameters $\gamma$, using
$-0.1$, $-0.01$, $0$, $0.01$, and $0.1$. The optimal set of parameters
for EL Embeddings are $embedding\_size = 50$ and $\gamma = -0.1$.  For
TransE (plain) $embedding\_size = 50$ (human) and $100$ (yeast), for
TransE (RDF) $embedding\_size = 400$ (human) and $200$
(yeast)\footnote{Detailed results are available at
  \url{https://www.dropbox.com/s/wresfh9fkfah4ei/supplement.pdf?dl=0}.}.
We report results on our testing set in Table \ref{tbl:results-yeast}
for the yeast PPI dataset and in Table \ref{tbl:results-human} for the
human PPI dataset.

For a query interaction $interacts(P_1, P_2)$ we predict interactions
of $P_1$ to all proteins from our training set and identify the rank
of $P_2$. Then we compute the mean of ranks for all interactions in
our testing set. We refer to this result as {\em raw mean rank}. Since
the interactions from our training and validation set will rank higher
than those in our testing set, we perform this evaluation excluding
training and validation interactions and report them as {\em
  Filtered}. We further report the area under the ROC curve (AUC)
which is more commonly used for evaluating PPI predictions
\cite{Kulmanov2017,neurosym}.

\section{Discussion}
Knowledge graph embeddings and other forms of relational learning
methods are increasingly applied in scientific tasks such as
prediction of protein--protein interactions, gene--disease
associations, or drug targets. Our work on EL Embeddings is motivated
by the need to incorporate background knowledge into machine learning
tasks. This need exists in particular in scientific domains in which
large formal knowledge bases have been created that can be utilized to
constrain optimization or improve search. Our embeddings are based on
the Description Logic \el which is widely used in life science
ontologies \cite{Smith2007} and combined with biological knowledge
graphs \cite{neurosym}.
EL Embeddings generate models for \el theories and do not rely on
precomputing deductive closures of graphs, and account for the
semantics of conjunction, existential quantifiers, and the bottom
concept (and therefore basic disjointness between classes).

While EL embeddings do not require negatives, we implement a form of
negative sampling by randomly changing classes in axioms, similarly to
how TransE and other knowledge graph embeddings generate negatives
\cite{Wang2017}. In future work, we intend to explore more of the
negatives that arise from the \el theories directly. For example, we
can encode the unique names assumption by asserting $\{ a \} \sqcap \{
b \} \sqsubseteq \bot$ for all instances $a,b \in I$, and infer
further negatives by exploring disjointness.

Another limitation of our method is the use of TransE as relational
model which does not allow us to capture role inclusion axioms or
model relations that are not one-to-one relations. Most of the EL
Embedding loss functions require no or little changes when using a
different relation model; however, as a consequence of using TransE as
relation model, for example the loss function for $\exists R.C
\sqsubseteq \bot$ is degenerate and will need to be
modified. Extending the relation model is not the only extension
possible to our model; in the future, we also intend to explore
improvements towards covering more expressive logics than \el.

\clearpage

\bibliographystyle{named}
\bibliography{ijcai19}

\end{document}